\documentclass[journal,11pt,onecolumn]{IEEEtran}
\usepackage[utf8]{inputenc}
\usepackage{stmaryrd}

\author{Mehdi~Bennis  and Salem~Lahlou
  \thanks{Mehdi Bennis is with the Centre of Wireless Communications, University of Oulu, Oulu, 90014, Finland (e-mail: mehdi.bennis@oulu.fi).}
    \thanks{Salem Lahlou is with the Mohamed bin Zayed University of Artificial Intelligence, Abu-Dhabi, UAE (e-mail: salem.lahlou@mbzuai.ac.ae).}
}

\usepackage[T1]{fontenc}
\usepackage{boisik}
\usepackage{lmodern} 

\usepackage{dsfont,enumerate,enumitem,comment,url}
\usepackage{multirow,booktabs,bigstrut,longtable,array}
\usepackage{colortbl,pdflscape,tabu,threeparttable,threeparttablex}
\usepackage[normalem]{ulem}
\usepackage{graphicx,color,rotating,soul}
\usepackage{makecell,xcolor,pbox,framed}
\usepackage{wrapfig,setspace,subfigure,float}
\usepackage{last page}
\usepackage[most]{tcolorbox}    
\tcbuselibrary{breakable} 

\usepackage{fdsymbol}
\usepackage{hyperref}

\colorlet{shadecolor}{gray!20}
\usepackage{avant}

\title{Semantic Communication meets System 2 ML: How Abstraction, Compositionality and Emergent Languages Shape Intelligence}

\begin{document}
\maketitle

\DeclareRobustCommand{\hlcyan}[1]{{\sethlcolor{green}\hl{#1}}}
\DeclareRobustCommand{\hlblue}[1]{{\sethlcolor{cyan}\hl{#1}}}

\begin{abstract}
\begin{tcolorbox}[colback=yellow!10, colframe=white!20!black, boxrule=1.9pt, boxsep=2pt,left=2pt,right=2pt,top=2pt,bottom=2pt]
    \noindent The trajectories of 6G and AI are set for a creative collision. However, current visions for 6G remain largely incremental evolutions of 5G, while progress in AI is hampered by brittle, data-hungry models that lack robust reasoning capabilities. This paper argues for a foundational paradigm shift, moving beyond the purely technical level of communication toward systems capable of semantic understanding and effective, goal-oriented interaction. We propose a unified research vision rooted in the principles of System 2 cognition, built upon three pillars: \textbf{Abstraction}, enabling agents to learn meaningful world models from raw sensorimotor data; \textbf{Compositionality}, providing the algebraic tools to combine learned concepts and subsystems; and \textbf{Emergent Communication}, allowing intelligent agents to create their own adaptive and grounded languages. By integrating these principles, we lay the groundwork for truly intelligent systems that can reason, adapt, and collaborate, unifying advances in wireless communications, machine learning, and robotics under a single coherent framework.

    \end{tcolorbox}

\end{abstract}

\tableofcontents
\newpage
  
\section{The  Wireless Communication World}
The wireless communication landscape is on the verge of a paradigm shift. As we look towards 6G (the sixth generation of wireless technology) we are seeing ambitious claims about connecting everything and everyone. But beneath these claims lies a more fundamental question: Are we approaching communication from the right perspective?

6G \cite{saad2019vision} promises to bridge the digital, physical, and biological worlds truly ushering in an era in which everything is sensed, connected, and intelligent. Yet the current 6G vision(s) of ``the more bits/s, more bandwidth, more base stations the better” or ``X times 5G requirements'' (e.g., in packet-error rate) are inadequate and not sustainable. On the AI/ML front, despite significant advances in the field, current solutions are brittle, energy-hungry, fail to generalize and are curve-fitting at best. Other visions such as O-RAN or statements like ``6G is whatever we will have in 2030'' are either stemming from a business standpoint or merely lacking ambition. In short, \textbf{6G today is nothing but an incremental evolution of 5G}.  These incremental advances may be rooted in two aspects: (i) the traditional radio engineering approach of the \textbf{``higher the better, the bigger the better''}, which is clearly not sustainable/scalable; (ii) radio engineers tend to cling to their domain knowledge instead of exploring the bigger vision with outside-of-the-box thinking. Instead of clinging to this path, could we flip this paradigm upside down by asking a different set of questions? Among these, can we transmit less data by instead leveraging reasoning and knowledge, thereby relaxing some of the stringent requirements? 
If 6G is genuinely about human-machine interaction/collaboration and communication for machines, can we learn from human's two modes of cognition \textbf{(System 1 and System 2)} and imbue this knowledge into machines to augment and assist us? System 1 represents fast, intuitive, and automatic cognitive processes, while System 2 involves slower, deliberate, and analytical reasoning \cite{kahneman2011thinking} (see Appendix~\ref{sec:system12} for details). By integrating both cognitive approaches into our communication paradigms, we will create systems that are both efficient and capable of deep understanding. These questions are worth delving into to unlock the true 6G revolution. And, as a matter of fact, this is the beginning of the revolution.

 \begin{wrapfigure}{R}{9.5cm}\centering \vspace{-1pt}
	\includegraphics[width=9.5cm]{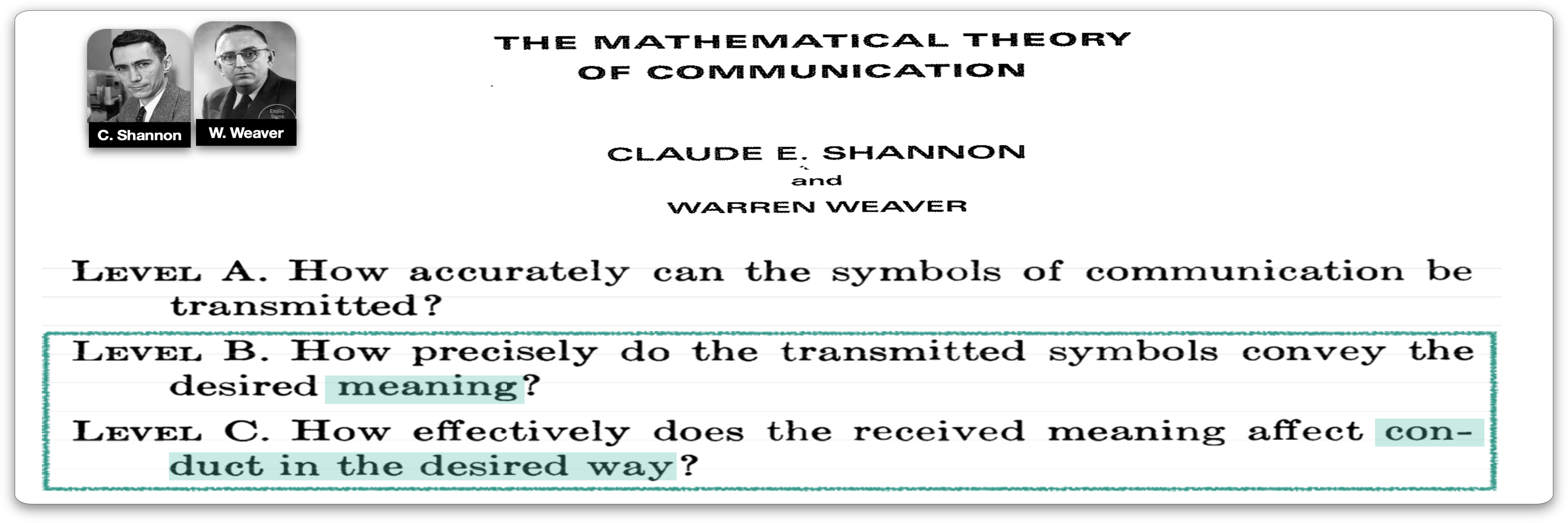} 
	\caption{\footnotesize Shannon's three levels of communication \cite{shannon1948mathematical}.}\vspace{-3pt}
	\label{shannonfig}
\end{wrapfigure}

The answer lies in the creative collision of two technological revolutions: Despite the tremendous progress made in communication, we are still in its infancy. Principles of communication theory can be traced back to the seminal work of Claude Shannon whose underlying communication problem (also referred to as the \textbf{technical problem}) was to reliably convey information from a sender to a receiver \cite{shannon1948mathematical}. Under this definition, communication is tantamount to reproducing at one point either exactly or approximately a message at another point (see  Fig. \ref{shannonfig}). Under level A, Shannon information is a statistical/syntactical description of information, concerned with the probability of co-occurrence of messages/symbols. Moreover, decoding a message requires a known mapping and an external observer (decoder) which is completely decoupled from context and (external) symbols are chosen by the external observer. This is clearly in contrast to how the brain and perceptual/biological systems operate. Although explicitly mentioned in the seminal paper (and Shannon himself issuing a word of caution in 1952 about making Shannon information universal in the bandwagon article \cite{1056774}), the \textbf{semantic and effectiveness} problems were irrelevant and put aside. In the age of robotics, LLMs and generative AI, it is now time to dust them off! 
While communication engineering has primarily focused on level A, the emerging challenges of 6G and human-machine collaboration demand that we now address the semantic and effectiveness problems  of communication. Moreover, recent research suggests that standard training methods in AI like reinforcement learning and imitation learning can lead to catastrophic risks through goal misspecification and emergent self-preservation behaviors \cite{bengio2025superintelligent}. Addressing these risks requires a fundamental rethinking of how we approach both AI and communication systems.

This evolution from purely statistical information processing towards semantic and effective communication represents a fundamental shift in how we conceptualize and optimize these increasingly complex wireless systems. In the following section, we explore different notions of information pertaining to these distinct communication levels.

\section{Information: The Semantic Chamaeleon}  
\label{chamaeleon}
 \begin{wrapfigure}{R}{10.5cm}\centering \vspace{-2pt}
	\includegraphics[width=10.5cm]{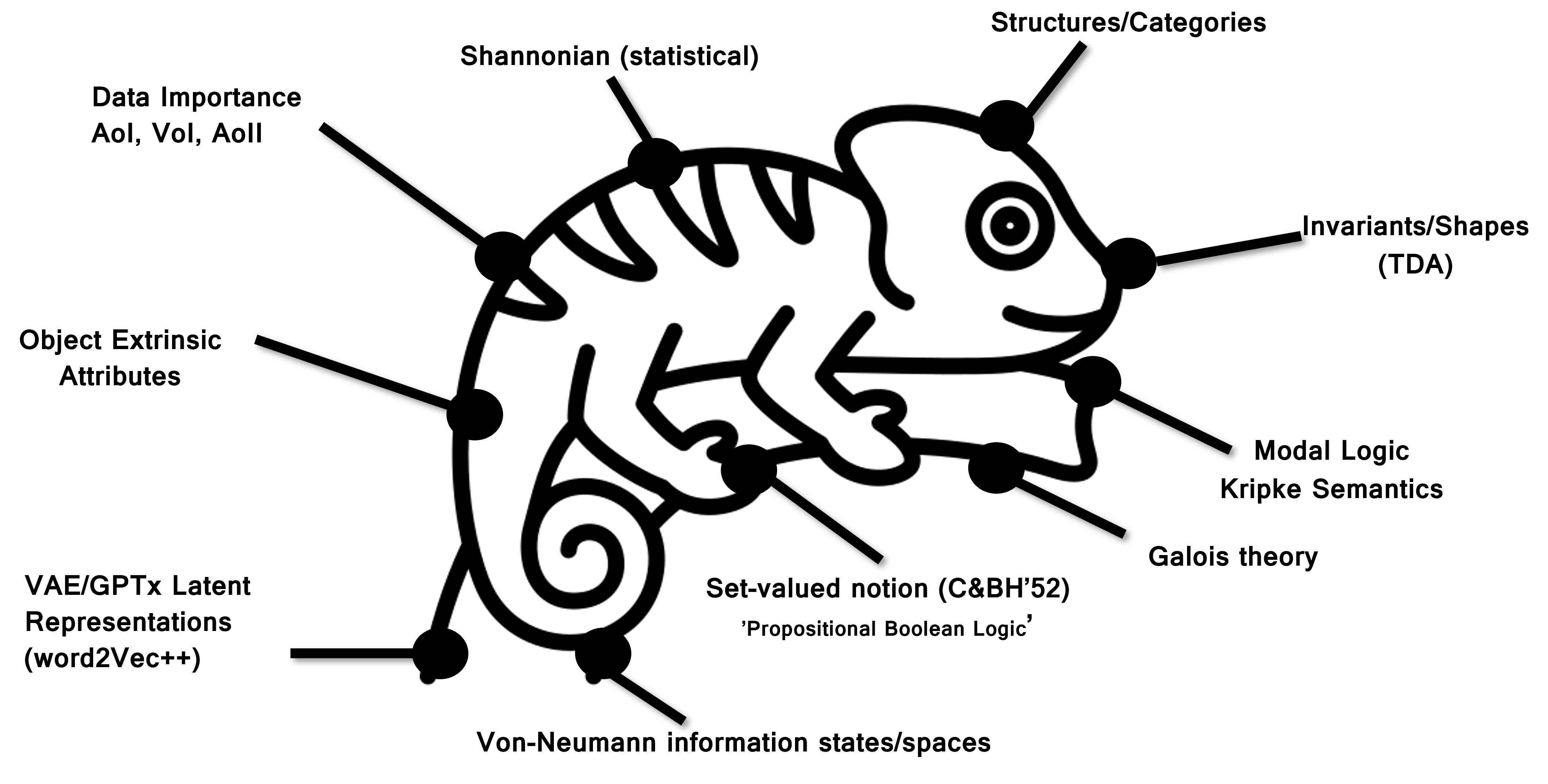} 
	\caption{\footnotesize Various shades of information.}\vspace{-7pt}
	\label{fig2}
\end{wrapfigure}
One of the key distinctions between Shannon's three levels of communication pertains to the notion of information. In contrast to  Shannon information (a scalar value associated with a probability distribution and measured by entropy), semantic information is concerned with information structures, shapes, spaces or more formally, information categories. Paraphrasing the French Mathematician René Thom,  ``the word  information should be replaced by the word \textsc{shape}'' \cite{Rossi2011RenTF}, underscoring the fact that information is first and foremost of topological nature \cite{baudot2019topological} (see Appendix~\ref{sec:topoinfo} for an accessible introduction to topological information theory). This perspective transforms how we conceptualize information in communication systems.

Depending on the mathematical language and algebraic structure of interest, different notions of information are at stake.  One of the earliest theories of information\textemdash known as the \textit{theory of ambiguity}\textemdash goes back to Galois theory, which associates a group to an algebraic equation. Galois theory offers a framework for understanding how different solutions to an equation are related, effectively capturing the structure of ambiguity or uncertainty. The associated group acts by permuting the set of solutions, characterizing how solutions may be indistinguishable from one another. As the group's size increases, so does the degree of indistinguishability among solutions.  Galois also examined the effect of conditioning on the group, in which acquiring additional information (represented by new numbers) restricts the group to a subgroup of the original one. In this algebraic setting, random variables $X$ and $Y$ are replaced by geometric spaces\textemdash sets on which a group acts, in which the relevant object is the quotient
$IG(X; Y) = \frac{G(X) \times G(Y)}{G(X,Y)}$, 
formed from the Cartesian product of $G(X)$ and $G(Y)$ divided by the Galois group of the joint extension $G(X,Y)$. This construction serves as an algebraic analogue to the probabilistic mutual information
$I(X, Y) = H(X) + H(Y) - H(X,Y).$

A different notion of information  relates to the invariant properties of topological spaces that are preserved under deformations. Specifically, homology theory of topological spaces associates algebraic structures (for e.g., homology groups) to topological spaces where these structures capture certain properties of the spaces that are preserved under continuous deformations.  The complexity of a topological space is given by the sum of dimensions of cohomological groups (Betti numbers
\footnote{Betti numbers describe topological features: $\beta_0$ counts connected components (islands), $\beta_1$ counts tunnels or holes (like in a donut), and $\beta_2$ counts cavities or enclosed voids (like the inside of a hollow sphere). These numbers help quantify the complexity of spatial structures.}). Information cohomology was also defined as an invariant associated with sheaves of modules over a category of statistical variables \cite{e17053253}.    Beyond algebraic and topological perspectives on information, and before the birth of Shannon information theory, Von Neumann’s notion of information states emphasizes uncertainty in decision-making within sequential games, specifically highlighting epistemic uncertainty, the knowledge an agent possesses about the system. In sharp contrast to Shannon information, these information states pertain to sets of possible worlds, where each world represents a state of the system. Within the logical realm, in $1952$  Carnap $\&$ Bar-Hillel proposed an  axiomatic notion of set-valued information based on a propositional Boolean language  \cite{685989}. 
Furthermore, in epistemic logic, the meaning of a logical formula (Kripke semantics) is defined by a set of possible worlds along with a truth assignment specifying the truth value of each formula in each world \cite{blackburn2001modal}. This formal framework enables the representation of what agents consider \textit{possible} or \textit{impossible}, building a structured model of knowledge states. It further supports reasoning about the truth of formulas across different worlds, allowing  modeling agents' knowledge and the refinement of their understanding of possible worlds under uncertainty. A sensorimotor and enactive perspective views information as the laws and knowledge that govern sensory signals and their relationships to actions, as perceived by agents \cite{noe2004action}. This perspective inherently involves the notion of sensory flow, since a single observation cannot reveal a law; instead, laws emerge across a sequence of observations where either the agent acts or the environment changes. Through interaction with the environment, specifically via (causal) interventions to test hypothetical actions, agents learn the underlying laws and structures of sensory signals, reasoning about their intents, beliefs, and goals. Depending on their cognitive abilities, agents can predict both future sensory inputs (low-level observations) and the expected outcomes of actions given current sensory information (higher-level predictions). Crucially, these learned structures are not mere mappings from sensory inputs to actions; they are compositional, hierarchical, and contextual, possessing their own syntax and algebra (semantics).

In summary, \textbf{several notions of information} exist beyond Shannon's framework, with each type (algebraic, topological, epistemic, and sensorimotor) capturing different ways of structuring and understanding uncertainty and knowledge. While 	Galois theory leads to understanding uncertainty in algebraic terms, homology theory shows how topological spaces encode invariant information that remains unchanged under deformations. Moreover, Von Neumann’s information states and epistemic logic focus on uncertainty about the system or knowledge states, whereas sensorimotor and enactive views of information focus on embodied knowledge and grounding information in real-world interactions.

\section{The AI/ML World}

Modern machine learning (ML) systems, often likened to Kahneman's ``System 1'' \cite{kahneman2011thinking} (see Appendix~\ref{sec:system12}), have achieved remarkable success in narrow tasks. These systems excel at rapid, intuitive pattern recognition but remain brittle, data-hungry, and prone to failures outside their training distribution. They lack interpretability, commonsense reasoning, and compositional generalization\footnote{Current deep learning models are not inherently compositional or strongly typed; their compositionality is syntactic (e.g., combining tokens) rather than semantic (e.g., reasoning about meaning) \cite{marcus2018deep}. Recent approaches for studying compositionality can be found in \cite{elmoznino2024complexitybasedtheorycompositionality}, \cite{lee2024geometricsignaturescompositionalitylanguage}.}. Furthermore, their massive computational demands raise sustainability concerns \cite{schwartz2020green}. To address these limitations, researchers increasingly turn to insights from computational neuroscience, metacognition, and cognitive psychology. Kahneman's seminal dichotomy between ``System 1'' (fast, automatic, heuristic-driven processing) and ``System 2'' (slow, deliberate, simulation-based reasoning) \cite{kahneman2011thinking} offers a framework for rethinking ML. While System 1-like models (e.g., deep neural networks) operate as black-box function approximators, System 2-like architectures aim to emulate human-like deliberation, constructing internal world models, running counterfactual simulations, and exhibiting curiosity-driven exploration \cite{buckner2024deep}. Crucially, these systems need not be mutually exclusive: a synergistic integration could allow System 2 to supervise and refine System 1, enabling generalization to novel environments, handling long-tail scenarios, and mitigating data scarcity \cite{garcez2020neurosymbolic}.
\begin{figure}\centering \vspace{-1pt}
	\includegraphics[width=10.5cm]{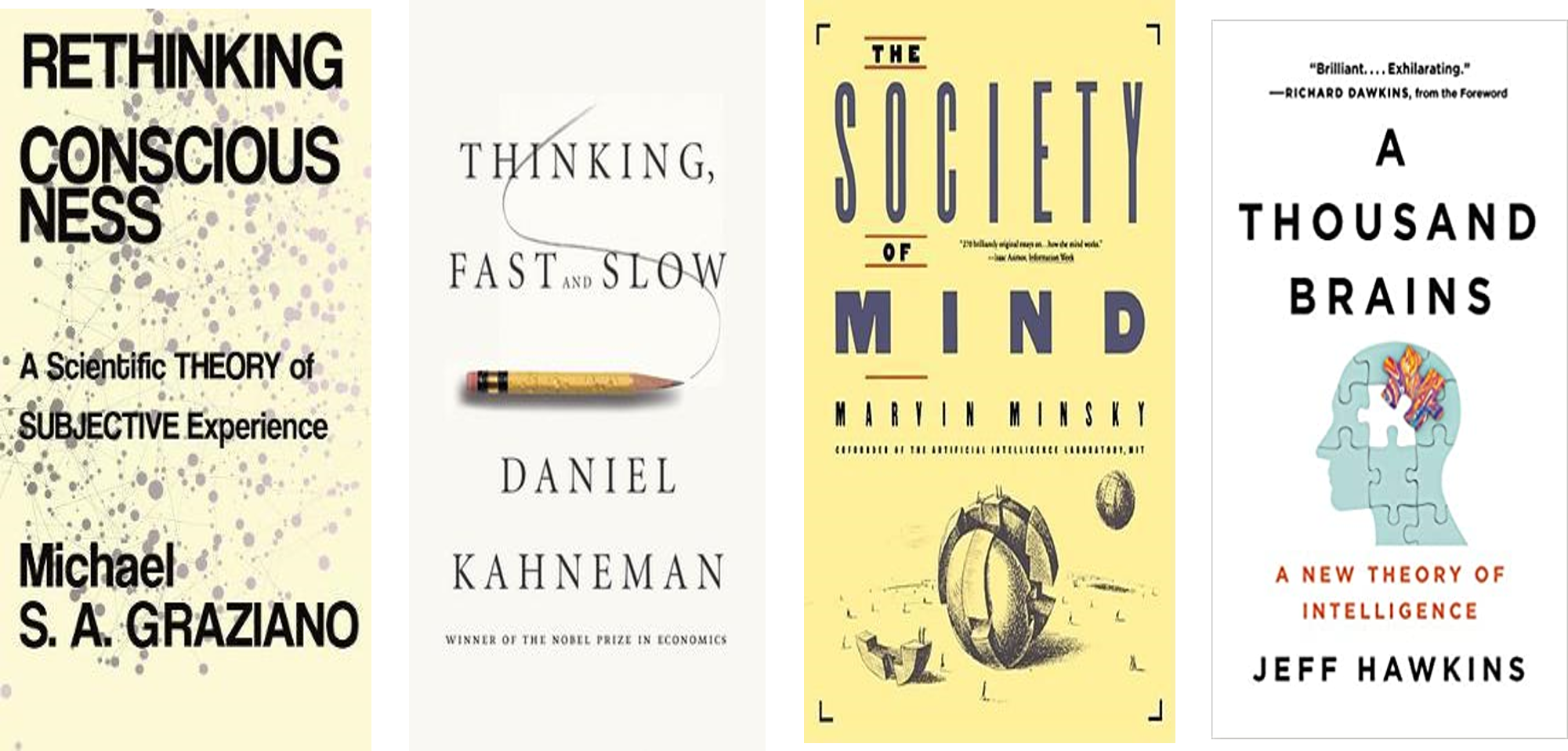} 
	\caption{\footnotesize  Daniel Kahneman (System 1 vs. System 2) \cite{kahneman2011thinking}, Michael S.A. Graziano (world model and metacognition), Marvin Minsky (emergence) \cite{minsky1986society} and Jeff Hawkins (a thousand brains hypothesis) \cite{hawkins2021thousand}. }
	\label{fig3}\vspace{-3pt}
\end{figure}

One of the key ingredients of System 2 ML is the ability to learn  concepts and abstractions from  (multimodal) sensorimotor signals via grounded interaction, essentially extracting meaningful patterns from raw perceptual data. Abstraction requires agents (or networks of agents) that sense their environment, maintain internal representations, and continuously update their beliefs based on new information. In these statistical-based learning approaches, agents actively infer an internal compressed representation (abstraction) of their external environment via internal feedback. This can be done using various techniques ranging from the free energy principle (FEP) \cite{FRISTON200670},  which models the action-perception loop in terms of variational (approximate) Bayesian inference, to self-consistent compressive control loops\footnote{Yi Ma's vision is rooted in learning minimal yet sufficient internal statistical models of   sensory inputs and errors are calculated in latent space. } \cite{ma2022principles}.  For modeling hierarchical abstractions (e.g., preferences or multi-resolution sensory data), lattices (ordered algebraic structures that organize concepts in hierarchical relationships) offer an expressive framework. These ordered algebraic structures establish equivalence relations and refinements between concepts, enabling both efficient inference and significant structural compression gains. Beyond these statistical approaches, from a formal standpoint and inspired from human's internal language of thoughts,    abstractions/concepts pertain to probabilistic programs, in which concepts are represented as simple probabilistic programs and richer concepts are built compositionally from simpler primitives \cite{doi:10.1126/science.aab3050}.  Human's internal language of thoughts is akin to a computer language, which encode and compress structures in various domains (math, music, shape, etc.).    This mental language can recursively compose primitives of number, space and repetition with variants.  Through this lens,   shape perception is tantamount to program inference, where the goal is to search for the minimal program from the space of all programs that captures the observed shape. This ability to abstract via compression is captured by the notion of minimum description length (MDL), whereby rather than leveraging the entire sequence length, task difficulty is proportional to  the MDL\footnote{Behavioral results
showed that the difficulty of memorizing a sequence was  modulated, not by the actual sequence length, but by the length of the program capable of generating it, whose complexity is measured in terms of MDL.}.
Yet another notion of abstraction (amiss in current AI systems) is analogy-making \cite{gentner2011computational}, which introduces a fundamentally different notion of similarity, beyond simple distance metrics. Unlike standard notions of similarity defined mathematically in a metric space (two sensory signals are similar if their difference is small) and distance is defined on the space of signals, there is another notion of similarity (for e.g., a heart is a pump), where similarity is not implied in any metric sense, but about how objects (heart and pump) look like in some representation space and  the way they interact with other things (the heart acts on blood in the same way the pump acts on a fluid).

\vspace{-1mm}
\subsection{World Models and Inference for Reasoning}
\label{lo}
Besides abstraction, another key ingredient of System 2 ML is world models for reasoning and planning (see Appendix~\ref{sec:worldmodel} for a detailed framework). Reasoning, at its core, requires two key components: a world model that encodes knowledge about the environment, and an inference machine that generates solutions or answers to queries based on this knowledge. This dichotomy is well-established in cognitive science and AI, where the world model captures causal relationships, uncertainty, and reusable abstractions, while the inference machine performs efficient search over combinatorial solution spaces \cite{lake2017building, bengio2017consciousness, bengio2025superintelligent}. World models serve as internal simulations of the environment that allow agents to predict outcomes and reason about hypothetical scenarios.

A world model (or distributions thereof) is based on the idea that an agent gets a state of the world (e.g., an image or video) and predicts the next state, either  resulting from its own action or the world itself. This allows planning and counterfactual reasoning about the system's future (if a particular failure were to occur now, what would be the best response?). We expand on more sophisticated notions of world models and their causal structure in Section~\ref{sec:desiderata}. World models have been extensively studied in model-predictive control, where  \textit{model} refers to the plant's state dynamics, and  in model-based RL, where a (world) model refers to  the state transition probabilities  used for look-ahead planning \cite{DBLP:journals/corr/abs-1803-10122}.    With the advent of big data and compute, planning and world models have  witnessed a resurgence in the AI community as evidenced by the joint embedding and predictive architecture (JEPA) \cite{Lecunvision}, latent state space models,  and many others. 
In particular, JEPA is a simple  approach for training a  world model  in a self-supervised learning (SSL) manner  using non-contrastive methods  (energy minimization).  Unlike  generative models (e.g., variational and masked autoencoders) that predict  object $Y$ (e.g., an image or video) with all details, including  irrelevant ones, JEPA  predicts  an abstract representation of $Y$. In a similar vein, drawing inspiration from  D. Kahneman and M. Graziano, Bengio et al. \cite{bengio2017consciousness} define a world model (or distribution thereof) as a factor graph where nodes/variables represent statements (true or false).  These  nodes/variables relate to each other in the factor graph via an energy/potential function (e.g., relating different statements). 

This fundamental separation of world model and inference machine has been explored across multiple domains. Generative flow networks (GFlowNets) offer a principled framework for decoupling the world model from the inference machine \cite{bengio2023gflownetfoundations}. By training an amortized inference machine to sample multimodal solutions consistent with a modular world model, GFlowNets address the overfitting and brittleness of monolithic architectures \cite{bengio2023scaling}. This separation enables combinatorial generalization, in which the inference machine leverages scaled deep networks for fast approximate inference, while the world model ensures solutions align with causal and uncertainty-aware abstractions \cite{bengio2023gflownetfoundations}. For instance, in drug discovery, GFlowNets sample molecular structures (inference) that satisfy constraints encoded in a chemical world model \cite{jain2022biological}.

It is worth mentioning that System 2 LLMs are currently under investigation such as using chain-of-thought approaches, Monte-Carlo tree search \cite{Kambhampati_2024} and many others. Recent work, such as \cite{bengio2023scaling}, has demonstrated that scaling alone is insufficient to resolve fundamental limitations like factual errors or unreliable reasoning in these systems \cite{bengio2025superintelligent}. While current approaches to System 2 LLMs show promise, they remain empirically driven and lack robustness. For instance, the ``inverse scaling'' phenomenon \cite{mckenzieinverse} demonstrates that larger models may counter-intuitively perform worse on tasks requiring structured reasoning. This highlights the need for inductive biases that explicitly model causality, uncertainty, and modular knowledge rather than relying solely on increased model size.

\section{Closing Loops Compositionally: Semantic Communication meets System 2 ML}

\begin{figure}[h!]\centering \vspace{-1pt}
	\includegraphics[width=10cm]{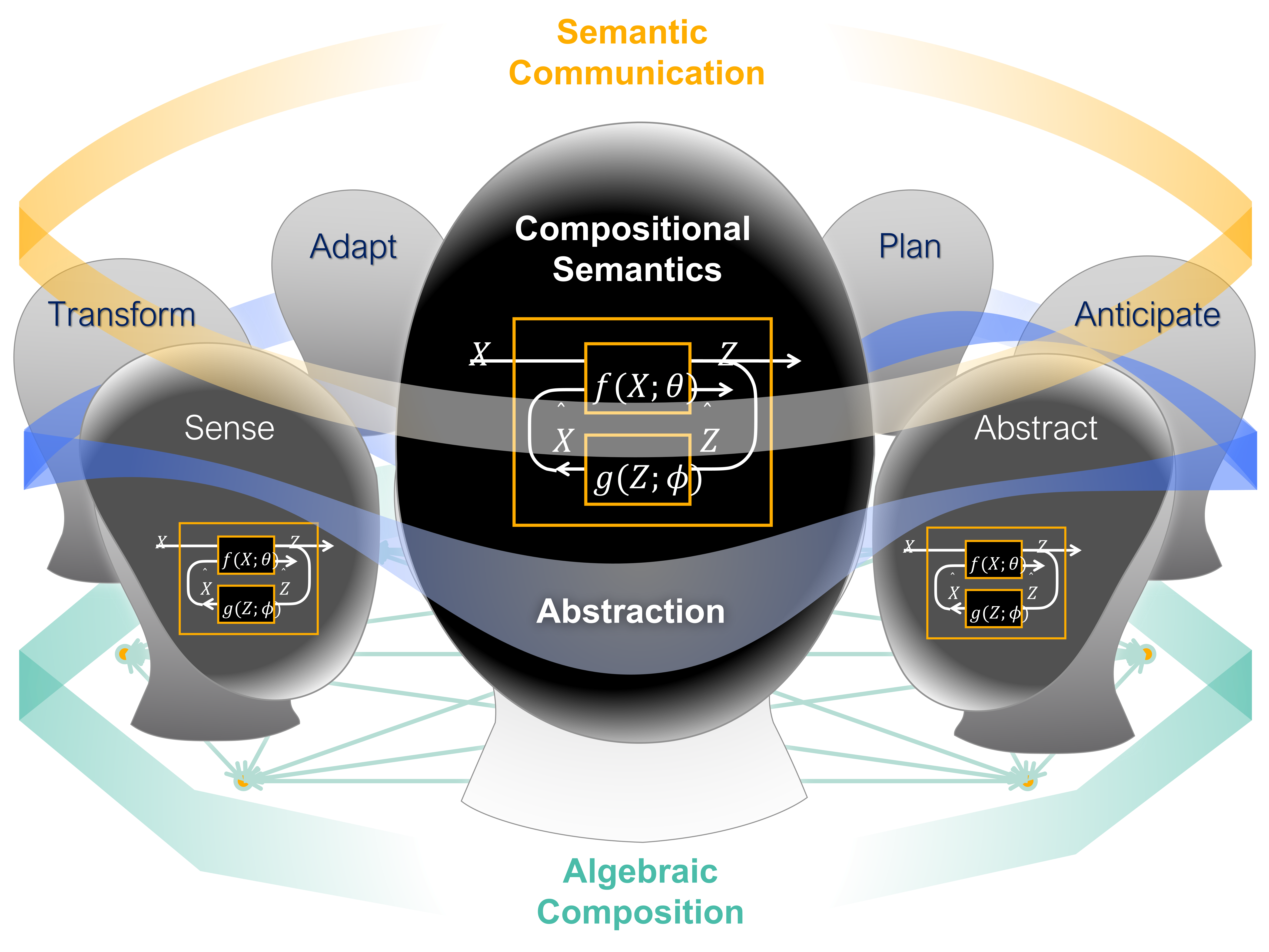} \centering	
\caption{\footnotesize  Confluence of abstraction, algebraic compositionality and semantic communication. Here, the semantics of active inference control-loops that sense, plan and adapt are composed. }
	\label{fig11}\vspace{-3pt}
\end{figure}

The confluence of semantic communication and System 2-type ML (see Appendix~\ref{sec:sc12} for a detailed comparison between System 1 SC and System 2 SC approaches) gives rise to a new research agenda for designing (truly) intelligent 6G sensing, communication, learning, control and reasoning systems. This integration represents a fundamental shift from traditional data-centric approaches towards knowledge-centric, reasoning-driven systems.  Akin to a society of minds, whether it is base stations, drones, sensors, vehicles/robots, an ML layer or agentic LLMs, this vision is about a collection of distributed, small and multimodal control-loops that sense, perceive, reason, plan, adapt and communicate to solve tasks that no single control-loop/agent can solve individually.  Under  this  unifying/integrative vision, agents depart from learning in raw data space (e.g.,  pixels, utterances, channels) towards learning semantic representations (abstractions and world models - see Appendix~\ref{sec:worldmodel}) of their environment via interaction. Akin to language, these representations are compositional, hierarchical, and contextual with an algebraic structure \cite{andreas2019measuring}. This structure enables the creation of extensive vocabularies that can be flexibly combined for counterfactual reasoning, planning, and communication, just as humans combine words and concepts to express unlimited meanings. Furthermore, from a communication perspective, the goal is to depart from reconstruction tasks (Shannon's level-A) towards agents  sensing, abstracting, planning and reasoning over sensory signals, intents, beliefs and goals. Instead of continuously transmitting raw data which may be either redundant, stale or of no value to a receiver, only the most important semantic information is learned, composed/transformed and communicated. 

Another ambitious line of research rooted in  abstraction and  compositionality is  how these distributed algebraic structures (different syntax, languages, models, priors, beliefs) communicate to solve a task. This mandates novel  emergent, adaptive and resilient  communication protocols \cite{lazaridou2020emergent} that arise naturally from agent interactions, in contrast to human-designed and hard-coded protocols that lack flexibility and generalizability. From an information standpoint, as underscored in Section \ref{chamaeleon}, these internal sensorimotor models (abstractions) give rise to different notions of information, not merely statistical (Shannon) information, but richer structures rooted in algebraic topology and logic (see Appendix~\ref{sec:topoinfo} for details). The ability to abstract and compose information,
concepts, subsystems and control loops provides a  calculus (or algebra)  that is instrumental in enabling \textbf{networks} that  not only reason and anticipate but also adapt to  disruptions  and unforeseen events and transform accordingly \cite{resil}. When it comes to System 2 ML/AI, compositionality will  emerge novel ML architectures going beyond current autoregressive LLMs (and their recent extensions), whereby an ML layer is equivalent to a control-loop or dynamical system and stacking layers corresponds to semantic compositionality (see Section V-B).  In addition, beyond reliability/robustness, resiliency has a significant impact on ML in terms of out of distribution (OOD) generalization, safety and explainability, notably in open-ended environments.
As AI systems are increasingly deployed in safety/mission-critical applications, relying on probabilistic approaches (e.g., model-free RL) without formal performance guarantees can lead to severe consequences. One way to ensure safety and trustworthiness is using formal verification methods and logic (temporal, epistemic, modal, etc.). These methods can help certify and explain the proper functioning of models, communication protocols and networks, as well as ascertaining resilience under belief manipulation and misleading information. Although solutions do exist such as using hard-coded and/or external human verifiers, or approaches that guarantee safety in expectation or asymptotically (e.g. safe RL type approaches),  more principled solutions providing strict formal guarantees  at all times are needed. These range from  internal verifiers  (as advocated in \cite{Kambhampati_2024}) to category-theoretic constructs embedding logical constraints into the (topos) structures, allowing to formally verify properties like consistency, correctness, and safety of communication protocols and models.   In particular, signal temporal logic (STL)-specifications provide a rich language to describe a resilient system in terms of: (i) recoverability (given an STL specification, a signal must recover from a violation within a pre-defined time period); (ii) durability (extent to which it can maintain its functionality for at least a pre-defined duration); and (iii) its quantitative semantics in terms of recoverability-durability pairs. The quantitative semantics of the logical specifications enable the incorporation of these requirements into an optimization framework for solving specific tasks \cite{chen2023stl, GirgisSTL}, providing formal guarantees in terms of soundness and completeness.

\vspace{+1mm}
Taken together, this vision promises transformative impacts across multiple domains: order-of-magnitude improvements in bandwidth-communication-energy efficiency, unprecedented adaptability to changing conditions, and network resilience  that maintains functionality even under significant disruption.
To make this vision a reality, several grand challenges cutting across several  disciplines will be investigated.

\subsection{Research Questions}

This research endeavor seeks to develop the theoretical and algorithmic
principles of \textbf{emergent and reasoning-driven compositional systems}.
Central to this vision are three core ingredients/pillars:  (1)  Abstraction: how agents construct meaningful representations from multimodal sensory data; (2) Compositionality: how these representations (abstractions) can be combined to form new meanings; and (3) Emergent Communication/Languages: how agents coordinate  and communicate through shared symbolic systems and learned communication protocols. The following \textbf{research questions} underpin some of the grand challenges associated with the  proposed vision:

       \begin{tcolorbox}[colback=white, colframe=blue!30!black, boxrule=.5pt,boxsep=3pt,left=3pt,right=3pt,top=3pt,bottom=3pt]
       \begin{wrapfigure}{R}{5cm}\centering \vspace{-2pt}
	\includegraphics[width=5cm]{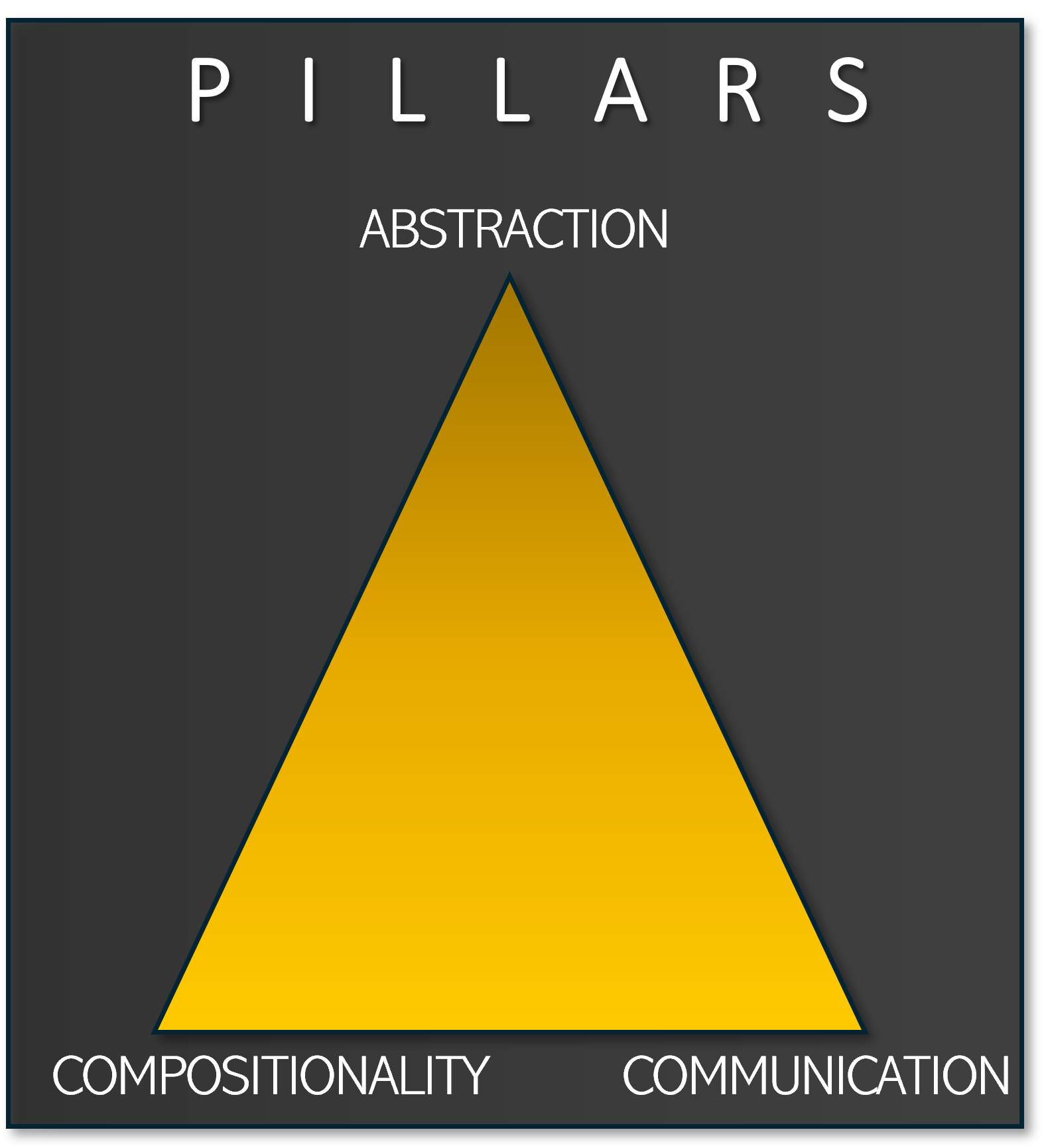} 
	\caption{\footnotesize Three Key Pillars underlying the proposed vision.}\vspace{-3pt}
	\label{fig5}
\end{wrapfigure}
      \underline{\textbf{Abstraction (Pillar 1)}}
                    \begin{itemize}
                        \item 
                    
            How can agents learn abstractions (equivalence classes, relations, partitions, programs, etc.) from their  multimodal sensorimotor signals through interactive experience?
           \item   How can agents  plan via their learned multiscale and multistep world models that balance detail and computational efficiency?
           \item How can epistemic logic frameworks (Kripke-type semantics) be leveraged to formalize reasoning about agents' knowledge states and enable more sophisticated planning?

           \end{itemize}
         \underline{\textbf{Compositionality (Pillar 2)}}
            \begin{itemize}
                           \item What formal algebraic frameworks best enable the composition of abstractions, concepts, and subsystems to create novel capabilities greater than the sum of their parts?
                           \item Under which conditions does system resilience emerge out of its individual components, and how does this resilience scale with network size, connectivity patterns, and topological structures?
                           \item How can we ensure that compositional systems maintain semantic coherence when combining elements from different domains or levels of abstraction?  
                           \end{itemize}
                                               
            \underline{\textbf{Emergent Communication $\&$  Languages (Pillar 3)}}

 \begin{itemize}

\item How  do distributed algebraic structures (different syntax, languages, models, priors/beliefs/knowledge) communicate to solve a task?
\item  What learning principles enable the emergence of adaptive and resilient  communication protocols between agents that plan and reason using their individual world models? 
\item What invariant properties characterize effective emergent languages across different domains and agent architectures?

\item  How can formal verification methods ascertain the functioning, interoperability, and explainability of these emergent models and communication protocols?
\end{itemize}

\end{tcolorbox}

\section{Key Desiderata}
\label{sec:desiderata}
\begin{wrapfigure}{R}{5cm}\centering 
\vspace{-2pt}
	\includegraphics[width=5cm]{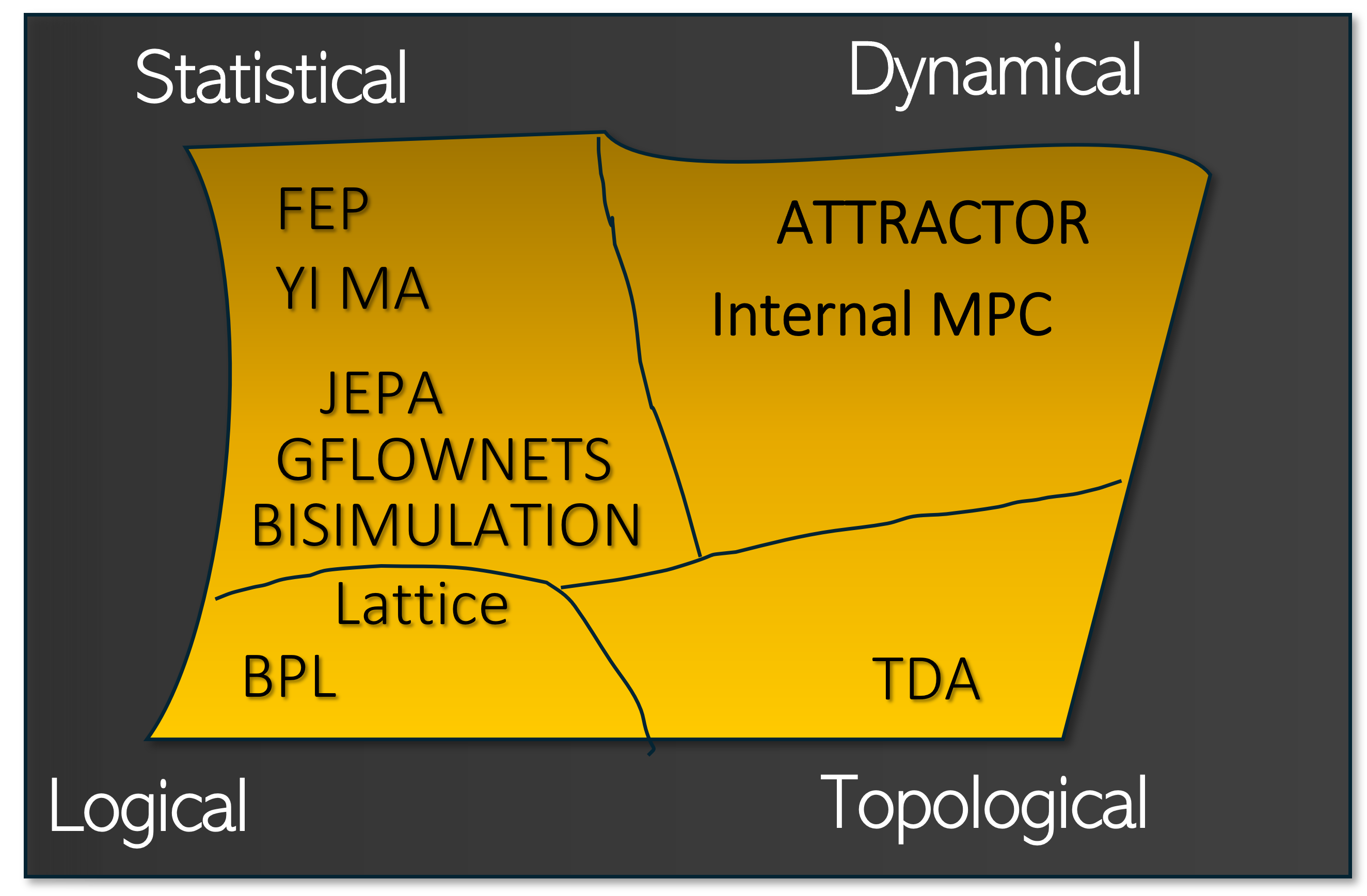} 
	\caption{\footnotesize Abstractions along the statistical, logical, dynamical and topological continuum.}\vspace{-3pt}
	\label{fig4}
\end{wrapfigure}
This section outlines the essential requirements and design principles for emergent reasoning-driven compositional systems. Building on the concepts introduced in previous sections, we elaborate on three interconnected pillars: abstraction, algebraic compositionality and  emergent communication/languages. These principles provide the foundation for our research agenda described in Appendix~\ref{sec:research_thrusts}.
\subsection{Abstraction, Anticipation, Adaptation  }

As mentioned in Section \ref{lo} and elaborated in Appendix~\ref{sec:worldmodel}, world models are a key building block for System 2-type ML. Nevertheless, current approaches to learning abstractions and world models (e.g., JEPA, FEP, GFlowNets, MPC, etc.) have significant limitations that must be addressed to realize our vision.
First, owing to  the diversity of real-world applications and  heterogeneous resource requirements, agents should  learn a collection/category of   sensorimotor world models (and their relations) through observations/interactions.  These world models should be
planning-compatible, causally structured,  multiscale, multistep and rapidly modifiable by counterfactual reasoning.   
Rather than simply predicting next states, sophisticated world models should capture sparse causal dependencies between relevant variables, allowing for more efficient representation and deeper reasoning capabilities.
World models across different modalities (RF\footnote{A wireless JEPA algorithm is proposed in \cite{10777043} for learning latent wireless dynamics from channel state information.}, images, video, LiDAR, etc.) may also come in different syntax (e.g., probabilistic, formal theories, etc,), calling for principled approaches for integrating them.   Moreover, agents should have an innate curiosity by continuously foraging for information based on their current models and uncertainty, generating hypothesis based on their world models (for e.g., using GFlowNets). Not only that, akin to prompt engineering,  besides world model and inference, agents need an attention policy to internally optimize/learn   what  information (or agent)  to attend to (query/prompt) for uncertainty reduction and maximum compression (minimum MDL).  For modeling hierarchical abstractions (e.g., preferences), lattices (ordered algebraic structures) offer an expressive and rigorous way, in terms of equivalence relations and refinements of sensory data and their representations, inducing fast inference and significant structural compression gains. From a formal logic standpoint, abstractions/concepts pertain to probabilistic programs, in which concepts are represented as simple probabilistic programs and richer concepts are built compositionally from simpler primitives \cite{doi:10.1126/science.aab3050}. Unlike statistical inference (e.g., FEP), in program inference, agents aim at  generating compressed mathematical theories (programs) of their external world, which entails searching for the minimal program from the space of all programs that explain  their sensorimotor data. Beyond statistical approaches to planning, from a formal logic standpoint,  planning is akin to automatic theorem-proving, where the agent's active inference block  proposes solutions to prove  theorems through a sequence of steps, and the world model sequentially checks the correctness of the proof. Another promising approach for modeling  abstractions and world models  is  via dynamical systems theory. Given a high-dimensional   non-linear dynamical system, the goal is to  learn to sample low-dimensional attractors  (discrete concepts)  from sensory data, for instance using GFlownets or Monte Carlo tree search (MCTS) or other techniques. Just like  language, these attractors (syntax) are   judiciously composed to solve various downstream tasks of interest.

\vspace{-3mm}
\subsection{Algebraic Compositionality } 

\begin{wrapfigure}{R}{7cm}\centering \vspace{-1pt}
	\includegraphics[width=7cm]{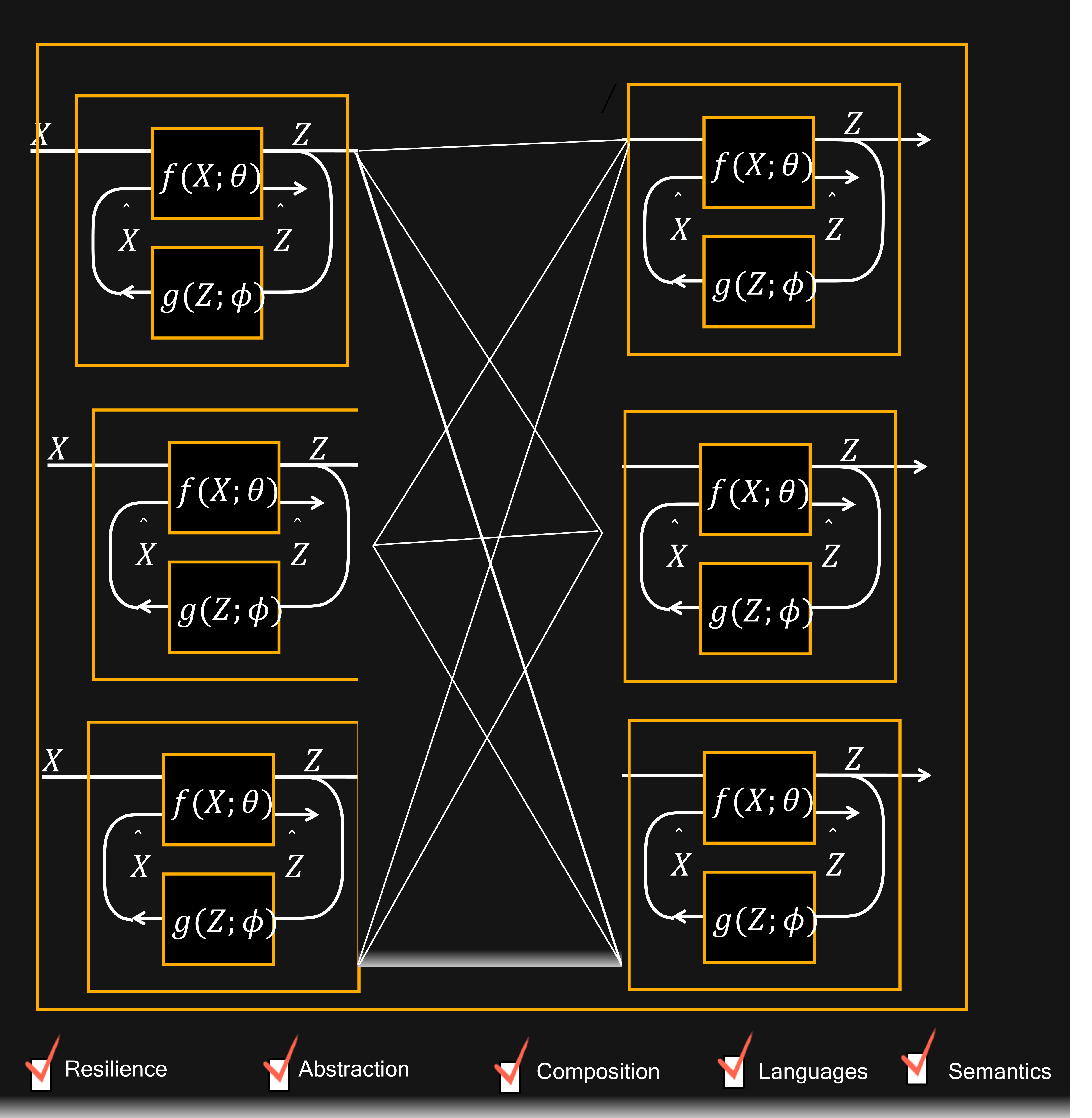} 
	\caption{\footnotesize Algebraic composition of the semantics of active inference agents. } \vspace{-3mm}
	\label{compose}
\end{wrapfigure} 
Humans have the remarkable cognitive capacity to form high-level  relational representations (abstractions) and compose them for rapid adaptation and generalization in changing environments. This ability to algebraically compose concepts, subsystems and dynamical systems/control loops is instrumental for reasoning, planning, communication and control. 
Central to this  goal is a novel   distributed sheaf-theoretic framework\footnote{Intuitively, sheaf theory provides a mathematical framework for answering the question: ``How can we combine local pieces of information into a coherent global picture?'' It's analogous to assembling a jigsaw puzzle, where each piece must fit with its neighbors, and the complete picture emerges only when all pieces are correctly arranged.} for learning multimodal representations and
their algebraic compositionality via first principles. Foundations of Sheaf theory  are leveraged to compose heterogeneous streams of semantic information
(abstractions) encoded in various algebraic data structures (vector spaces, lattices and topological spaces).

Sheaf theory \cite{bredon_sheaf_1997} provides a mathematical framework for the ``local to global'' problem: how to integrate locally consistent information into globally coherent structures. It encodes assignments of data/models to geometric or topological structures and provides principled ways to ``glue'' compatible local representations into unified \textit{global views}. This makes it ideal for composing heterogeneous information across modalities and agents. Owing
to its algebraic topological nature, sheaf theory allows to study such data/model relationships exploiting
symmetries and transformations. The core idea is that, instead of studying the space itself, we study algebraic
data structures (i.e., sheaves) and their transformations. This describes the space of semantic information
where reasoning tasks (data transformations/fusion, join, meets, products/coproducts and queries) occur. Beyond consensus, sheaf theory can also help solve disagreements
when agents with different world models (abstractions and beliefs
over their world) communicate and coordinate, by aligning their
beliefs via interaction to better predict each others actions and
intents.  
Furthermore, when it comes to ML, a sheaf-theoretical formulation can give rise to novel ML architectures, in which each ML layer corresponds to a control-loop or dynamical system and stacking layers corresponds to glueing/composing their semantics.

\begin{wrapfigure}{}{7.5cm}\centering 
	\includegraphics[width=7cm]{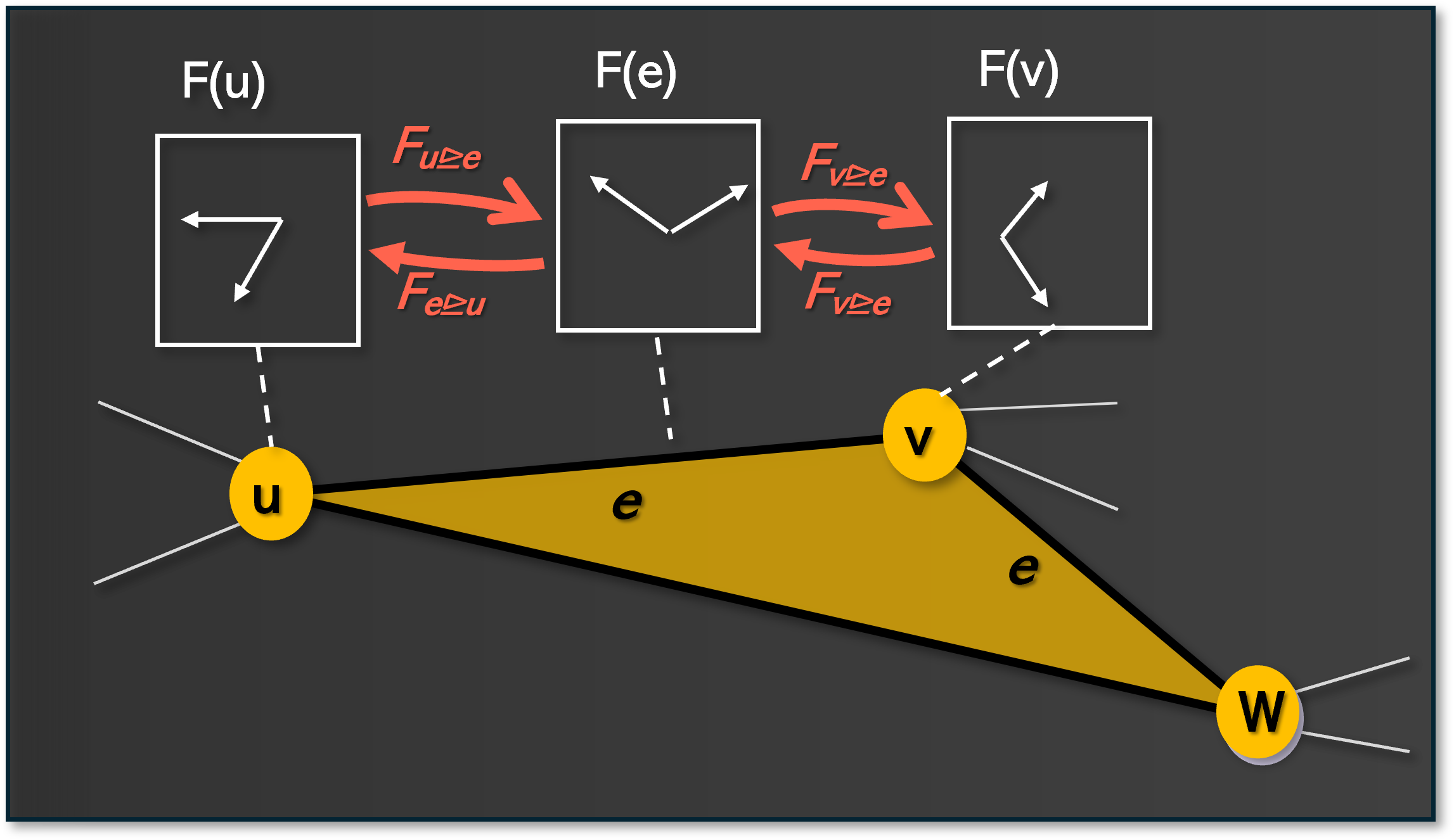} 
	\caption{\footnotesize  Sheaf-theoretic approach to communication: Vector space-valued Sheaves (data structures) on edge (e) and vertices (u,v,w) with learnable (linear) restriction maps \cite{issaid2025tacklingfeaturesampleheterogeneity}.} 
	\label{sheaves}
\end{wrapfigure}
When it comes to communication, algebraic compositionality  plays a central role. At its core, distributed algebraic structures (i.e., different syntax, language, structures and models) need to communicate  and coordinate to achieve a goal (for e.g., multimodal sensory integration or compositional reasoning). Algebraic compositionality is also a key enabler for semantic communication among distributed agents, whereby communication is tantamount to agents \textit{composing} their internal information spaces (sheaf of world models)  for mutual predictability; a prerequisite for understanding. Going beyond classical (i.e., probabilistic) FEP \cite{sakcak2023mathematicalcharacterizationminimallysufficient, WAFR}, each agent can be cast as  an information transition system (ITS),  and local information spaces (semantics) are composed, for example, via sheaf diffusion, to  solve collaborative tasks.  Going further,  a rigorous  framework for compositional multi-agent systems is through the lens of category theory (CT), which captures the mathematical essence of composition (process by which many parts make  a whole), allowing to build multiscale nested systems and reasoning about them.

\vspace{-3mm}

\subsection{Emergent Communication and Languages}

The proposed vision emphasizes the ability of agents to develop new communication languages and protocols based on their multimodal sensory data and knowledge, while being grounded  in interactions. Specifically, the reasoning-driven semantic communication framework enables agents to learn communication strategies by leveraging topological/logical/categorical abstractions and reasoning capabilities. This research  explores how communication emerges among agents with partial information and limited knowledge, ensuring robustness to different priors, beliefs, and structures. 

   Current approaches to learning communication protocols fail to generalize and lack robustness in unseen conditions and environments. To address this, we propose a cooperative partially-observable MDP (POMDP) setting, where agents learn abstractions (emergent equivalence classes) of their environment through interaction. Rather than learning complex policies in high-dimensional spaces, agents discover minimal yet sufficient models that capture only the essential information needed to solve tasks collaboratively. Rather than learning a state-action policy in high-dimensional observation/state spaces, concepts of bisimulation from automata theory can be used, which defines an equivalence relation between states that captures exact behavioral similarity in terms of reward. Two states are equivalent if their reward is the same and the transition probability in the abstract space is identical. Formally, an equivalence relation $B$ between states is a bisimulation relation, if for all states, $s_i$, $s_j \in \mathcal{S}$ that are  equivalent under $B$, (denoted $s_i\equiv_{B}s_j$)  $R_i(s_i,a)=R_j(s_j,a)$ and $P(G|s_i,a)=P(G|s_j,a)$ for $\forall a \in \mathcal{A}$ for $\forall G\in S_B$, where $S_B$ is the partition of $\mathcal{S}$ under relation $B$ (set of all groups $G$ of equivalent states), and  $P(G|s,a)=\sum_{s'\in G }P(s'|s,a)$; the latter means two states are equivalent if they have the same probability distribution into the same abstract states.       An abstract representation is learned  such that  the $L_1$ distance  between any two states is a measure of their bisimilarity.   Besides bisimulation, other  algebraic, topological and logical  abstractions can be used.  
   These continuum of abstractions will collectively form a library of  protocols that can be ranked, composed, and analyzed.

\begin{wrapfigure}{r}{8cm}\centering \vspace{-1pt}
	\includegraphics[width=8cm]{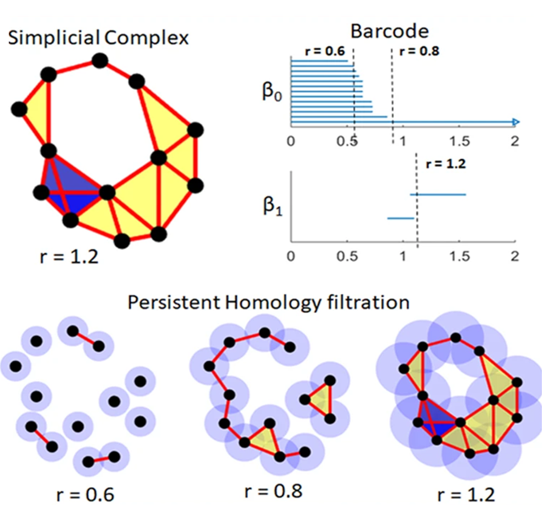} 
	\caption{\footnotesize  Simplicial complexes, persistent homological filtration and barcodes.} \vspace{-5pt}
	\label{tda}
\end{wrapfigure}

Additionally, analyzing the properties of these emergent languages (such as their invariants, topological structures, and compositionality) is essential. In terms of reasoning, agents will utilize multiscale, multistep world models for hierarchical planning under uncertainty. From a logical perspective, epistemic logic (Kripke-type) will support abstraction and reasoning, facilitating the development of adaptive, resilient, and verifiable semantic communication protocols, in contrast to purely statistical approaches that lack formal guarantees. Unlike existing state-of-the-art methods, this framework integrates both data- and reasoning-driven approaches to develop communication protocols tailored to specific application requirements and resource constraints, including energy efficiency, reliability, and signaling overhead. Beyond pairwise interactions, emergent communication protocols are also influenced by higher-order interactions, network topology, and their dynamics. These factors play a crucial role in how information is processed, structured, and how systems adapt and evolve over time. 
In this context, Topological Data Analysis (TDA) provides essential tools for understanding how the shape of data influences multi-agent networks. TDA reveals structural patterns beyond pairwise connections, capturing higher-order interactions that are crucial for system resilience and adaptability (see Appendix~\ref{sec:topoinfo} for mathematical foundations). Specifically, persistent homology reveals topological invariants (shapes) within data through a multi-resolution perspective of nested simplicial complexes. Key TDA metrics, such as Betti numbers, capture persistent geometric properties, tracking the birth and death of homology classes as a function of a varying parameter $r$ (see Fig. \ref{tda}). \hfill

  These topological invariants serve as abstract representations, structural semantics and resilience metric.

From the perspective of dynamical systems theory, these complex networks exhibit hierarchical and modular structures composed of coupled, nonlinear and high-dimensional dynamical systems. Understanding how higher-order interactions influence adaptation and resilience to  perturbations and potential failures is essential. 
A system loses resilience when it reaches a critical point and undergoes a bifurcation, i.e., a sudden transition to a different attractor state. Moreover, depending on the magnitude of a perturbation (small or large), higher-order interactions can enhance linear stability (resistance to minor disturbances) while simultaneously reducing the stability of basins of attraction (a global measure that determines system response to larger disturbances). 

Altogether, these desiderata inform our  research agenda detailed in Appendix~\ref{sec:research_thrusts}, which outlines specific research thrusts and proof-of-concept applications that demonstrate our vision in practice.

\bibliographystyle{ieeetr}  
\bibliography{IEEEabrv,vision} 
\renewcommand{\thesubsection}{\Alph{subsection}} 
\renewcommand{\thesubsubsection}{\thesubsection.\arabic{subsubsection}} 

\makeatletter
\renewcommand{\subsection}{\@startsection{subsection}{2}{\z@}%
  {1.5ex plus 1ex minus .2ex}%
  {1ex plus .2ex}%
  {\normalfont\large\bfseries}} 

\renewcommand{\subsubsection}{\@startsection{subsubsection}{3}{\z@}%
  {1.25ex plus 1ex minus .2ex}%
  {0.75ex plus .2ex}%
  {\normalfont\bfseries}} 
\makeatother

\section*{Acknowledgments}
The authors express their gratitude to Yoshua Bengio for his valuable insights and constructive feedback.

\appendix

\subsection{Fundamentals of System 1 and System 2 Cognition}\label{sec:system12}

Communication systems are increasingly designed to collaborate with and augment human capabilities. To achieve this effectively, understanding the dual nature of human cognition becomes essential. This appendix outlines the key principles of System 1 and System 2 cognition and their relevance to semantic communication.

\subsubsection{Dual-Process Theory of Cognition}

The dual-process theory, popularized by psychologist Daniel Kahneman in his book ``Thinking, Fast and Slow'' \cite{kahneman2011thinking}, distinguishes between two modes of thought:

\begin{itemize}
    \item \textbf{System 1} operates automatically, quickly, with little or no effort, and no sense of voluntary control. It is intuitive, associative, and pattern-based.
    
    \item \textbf{System 2} allocates attention to effortful mental activities, including complex computations, logical reasoning, and careful deliberation. It is slow, analytical, and resource-intensive.
\end{itemize}

These systems work in tandem but serve different purposes. System 1 allows us to navigate familiar situations with minimal cognitive load, while System 2 activates when we encounter novel problems that require careful analysis. A summary is provided in Table \ref{tab:systems}.

\begin{table}[h]
\centering
\begin{tabular}{|p{3.5cm}|p{5cm}|p{5cm}|}
\hline
\textbf{Characteristic} & \textbf{System 1} & \textbf{System 2} \\
\hline
Processing speed & Fast & Slow \\
\hline
Cognitive effort & Low & High \\
\hline
Operation & Automatic, unconscious & Controlled, conscious \\
\hline
Capacity & High capacity, parallel & Limited capacity, serial \\
\hline
Error susceptibility & Prone to biases and heuristics & Less prone to biases when actively engaged \\
\hline
Learning mechanism & Associative learning through repetition & Rule-based learning through instruction \\
\hline
Evolutionary origin & Ancient & More recent \\
\hline
Examples & Recognizing faces, navigating familiar routes, intuitive responses & Mathematical reasoning, logical deduction, planning complex tasks \\
\hline
\end{tabular}
\caption{System 1 vs. System 2 Cognitive Processes}
\label{tab:systems}
\end{table}

\subsubsection{Relevance to Communication Systems and AI}

Current communication systems and AI predominantly employ what could be considered System 1-like processing. The systems are pattern-based, statistical, and trained on large datasets to respond quickly to familiar inputs. However, they often lack the analytical reasoning capabilities of System 2 thinking.

For semantic communication and next-generation systems, the integration of both modes offers several advantages:

\begin{itemize}
    \item \textbf{Efficiency-Intelligence Balance}: System 1-like processes handle routine communication tasks efficiently, while System 2-like reasoning addresses complex, novel situations that require deeper understanding.
    
    \item \textbf{Adaptability}: A dual-system approach allows communication systems to adapt between fast, low-resource transmission for familiar contexts and more deliberate, semantic-rich communication when precision and understanding are crucial.
    
    \item \textbf{Contextual Awareness}: System 2-like reasoning enables machines to consider context, meaning, and consequences, which are essential aspects of semantic communication (Shannon's levels B and C).
    
    \item \textbf{Resource Optimization}: Like humans who conserve cognitive resources by relying on System 1 for routine tasks, communication systems can optimize resource allocation by engaging deeper reasoning only when necessary.
\end{itemize}

\subsubsection{Implementation Challenges}

Implementing true System 2-like capabilities in machines remains challenging. Current approaches include:

\begin{itemize}
    \item \textbf{Neuro-symbolic AI}: Combining neural networks (System 1-like pattern recognition) with symbolic reasoning (System 2-like logical processing) \cite{hitzler2022neuro,garcez2023neurosymbolic}.
    
    \item \textbf{Large Language Models with Chain-of-Thought}: Prompting models to generate step-by-step reasoning before arriving at conclusions \cite{wei2022chain,lahlou2024port}.
\end{itemize}

The development of communication systems that genuinely integrate both cognitive modes represents a frontier in both telecommunications and artificial intelligence research, with profound implications for how machines and humans will communicate in the 6G era and beyond.

\subsection{Topological Information: A Primer}\label{sec:topoinfo}

To understand semantic communication, it's valuable to explore how information can be represented as shapes and structures rather than just probabilities. This appendix provides an accessible introduction to topological approaches to information.

\subsubsection{From Numbers to Shapes}

Shannon's information theory quantifies information as a number (bits) derived from probability distributions. However, this approach discards the structural relationships within data, which are the very elements that often carry meaning. Topological information theory retains these structural relationships.

\subsubsection{Key Concepts}

\begin{itemize}
    \item \textbf{Topology}: The mathematical study of shapes and spaces that are preserved under continuous deformations (stretching, bending, but not tearing).
    
    \item \textbf{Homology}: A method to assign algebraic structures (like groups) to topological spaces, capturing essential features like holes, voids, and connectivity.
    
    \item \textbf{Betti Numbers}: Numerical invariants that count topological features:
    \begin{itemize}
        \item $\beta_0$: Number of connected components
        \item $\beta_1$: Number of one-dimensional holes (tunnels/circles)
        \item $\beta_2$: Number of two-dimensional voids (cavities)
    \end{itemize}
    
    \item \textbf{Persistent Homology}: Technique to analyze how topological features persist across different scales, revealing multi-scale structure in data.
\end{itemize}

\subsubsection{Applications to Communication}

Topological approaches to information offer several advantages for semantic communication systems:

\begin{itemize}
    \item \textbf{Robust Feature Extraction}: Topological features are invariant to many transformations, making them robust descriptors of data.
    
    \item \textbf{Multi-scale Analysis}: Capturing structures at different scales allows communication systems to adapt to different levels of detail.
    
    \item \textbf{Relational Information}: Preserving relationships between data points enables reasoning about context and meaning.
    
    \item \textbf{Dimensionality Reduction}: Topological summaries can compress complex data while retaining essential structural information.
    \item \textbf{Semantic Communication}:     persistence diagrams (PDs) as  topological signatures of raw point cloud data yield more effective use of transmission channels, enhanced degrees of freedom for incorporating error detection/correction capabilities, and improved robustness to channel imperfections \cite{asirimath2024rawdatastructuralsemantics}.  
\end{itemize}

\subsubsection{Example: Message Understanding}

Consider two communication systems:

\begin{itemize}
    \item A \textbf{Shannon-based system} might accurately transmit all the words in a message but miss the context that gives them meaning.
    
    \item A \textbf{Topology-based system} might identify key structural relationships in the message (which entities are related, how concepts cluster together, and what logical patterns connect ideas) preserving the semantic content even if some individual words change.
\end{itemize}

This topological view of information aligns with how humans process communication, focusing on structures and relationships rather than isolated symbols.

\subsection{World Model Framework and Compositional Inference}\label{sec:worldmodel}

This appendix provides a conceptual and technical overview of our proposed framework for implementing world models and compositional reasoning capabilities in AI systems. It expands on the ideas introduced in Section~\ref{lo} while focusing on accessibility.

\subsubsection{Core Components and Principles}

Our proposed framework integrates two essential elements: world models that represent knowledge about the environment, and inference machinery that reasons with this knowledge. This separation is inspired by cognitive science theories like the Global Workspace Theory \cite{dehaene2014consciousness}, which suggests that human cognition involves specialized modules communicating through a limited capacity workspace.

\textbf{World Model:} The world model encodes knowledge as a sparse dependency graph \cite{bengio2017consciousness}, where:
\begin{itemize}
    \item Each node represents a compositional statement about the world (e.g.,  ``the cup is on the table'')
    \item Relationships between statements are encoded as dependencies in the graph
    \item The structure is factorized, making inference tractable despite the complexity of real-world environments
\end{itemize}

\textbf{Inference Machinery:} To efficiently reason with this world model, we propose to employ Generative Flow Networks (GFlowNets) \cite{bengio2023gflownetfoundations}. GFlowNets are particularly well-suited for this task because:

\begin{itemize}
    \item They efficiently sample from complex, multimodal distributions
    \item They provide a principled way to explore spaces of possible solutions or explanations
    \item They decouple the world model (what is known) from the inference process (how to reason with that knowledge)
\end{itemize}

Unlike traditional deep learning approaches that often conflate knowledge and inference in a single model, this separation allows for more robust reasoning, especially in novel situations.

\subsubsection{Generative Flow Networks for Compositional Reasoning}

Generative Flow Networks (GFlowNets) \cite{bengio2023gflownetfoundations} are a relatively recent framework for learning to sample from complex probability distributions. Unlike discriminative models that map inputs to outputs, or generative models that produce data samples starting from a training dataset, GFlowNets learn to sample objects proportionally to a given reward function.

\textbf{Why GFlowNets for Reasoning?} GFlowNets offer several advantages that make them ideal for compositional reasoning:

\begin{itemize}
    \item \textbf{Structured exploration:} They efficiently explore combinatorial spaces by building solutions step-by-step
    
    \item \textbf{Diversity:} They naturally generate diverse solutions rather than converging to a single optimum
    
    \item \textbf{Compositional sampling:} They construct complex objects through sequential decisions, mirroring how humans build complex thoughts from simpler concepts
    
    \item \textbf{Uncertainty representation:} They can sample from a distribution of possible hypotheses, capturing uncertainty in reasoning
\end{itemize}

In our framework, we propose modular GFlowNets where each module focuses on specific aspects of reasoning (e.g., object identification, spatial relationships, action planning). These modules communicate through a shared workspace with limited capacity, creating a chain-of-thought process similar to human reasoning.

\subsubsection{Learning Meaningful Representations}

The framework learns to represent concepts as low-dimensional attractors in a dynamical system \cite{nam2023discrete}. These attractors function like symbols in a compositional language:

\begin{itemize}
    \item \textbf{Discrete concepts:} Each attractor corresponds to a fundamental concept (e.g.,  ``cup,''  ``table,''  ``on'')
    
    \item \textbf{Compositional meaning:} Complex meanings emerge from combining these basic units (e.g.,  ``cup on table'')
    
    \item \textbf{Grounded semantics:} The meaning of concepts is grounded in sensorimotor experience rather than arbitrary symbols
\end{itemize}

The learning process is formalized as Bayesian inference over the structure and parameters of the world model, with GFlowNets sampling from this posterior distribution. This allows the system to represent uncertainty and refine its beliefs with new evidence.

\subsubsection{Key Advantages}

This approach offers several advantages over traditional deep learning methods:

\begin{itemize}
    \item \textbf{Combinatorial generalization:} The ability to combine learned concepts in novel ways to solve new problems
    
    \item \textbf{Sample efficiency:} Structured reasoning reduces the need for extensive training data
    
    \item \textbf{Interpretability:} The compositional nature of the representations makes reasoning steps more transparent
    
    \item \textbf{Uncertainty awareness:} The system explicitly represents what it knows and doesn't know
    
    \item \textbf{Modularity:} Specialized modules can be combined to solve complex tasks, improving scalability
\end{itemize}

Recent work has demonstrated the potential of this approach in domains ranging from scientific discovery \cite{jain2022biological} to causal reasoning \cite{deleu2022bayesian}.
\subsection{System 1 SC vs. System 2 SC}\label{sec:sc12}

Despite the large body  of articles, the topic of semantic communication remains very fragmented in terms of what the word semantic means (beyond its reduction to Greek etymology), what semantic communication means or even what semantic information is to begin with? In fact, the overwhelming perception is that anything goes (from goal-oriented communication, task-oriented communication to semantics-empowered communication and Deep joint source and channel coding, to mention a few). Likewise, and as shown in Section \ref{chamaeleon}, semantic information  seems to include just about anything ranging from age of information (AoI) and variants, ML-based latent representations (e.g., VAE, GANs, etc.), set-valued notions of information (Carnap Bar-Hillel), topological invariants, truth-conditional semantics, and much more. In an effort to unify this critically important field of research with broad implications, and drawing inspiration from Kahneman’s System 1 vs. System 2 framework, \cite{panel} proposed, back in 2021, a distinction between System 1 SC and System 2 SC . This   articulates a distinction between current (System 1-type) approaches from the more difficult System 2-type approaches whose success hinges on leveraging different fields of mathematics (beyond statistics and probability theory). What is more? this two-system connotation is also meant to build a missing bridge towards the AI community, robotics and many others.

\subsubsection{System 1 SC: Current Approaches}

System 1 SC encapsulates all the current progress found in the literature (a recent book on the topic can be found in \cite{walidsemcom}). This boils down to applying blackbox ML to any communication problem across different OSI layers (e.g.,  physical and MAC layers). These statistical and data-hungry techniques excel at interpolation and reconstruction-type tasks, however they are brittle and fail to generalize or reason, to mention a few of their caveats. AoI and all its variants (including VoI, QoI and many other variants) are syntactic notions of information that are well-suited for Shannon's level-A reconstruction-type problems. However, this is not the type of information biological systems (with agency) use to plan, reason, survive and interact with other agents. Interestingly, Shannon himself warned us against using that notion of information in his 1952 paper. In the same vein, text-to-image transformations or other recent ML advances (DALL-E, CLIP, GP4+) are all System 1-type approaches despite their spectacular jaw-dropping performance. 

\textbf{Example 1: Image Transmission in System 1 SC.} Consider transmitting an image of a traffic scene from a roadside camera to a control center. In a System 1 SC approach, the system might:
\begin{itemize}
    \item Use a neural network to encode the image into a compressed latent representation
    \item Prioritize transmission of certain features based on learned patterns (e.g., moving objects over static background)
    \item Reconstruct the image at the receiver using a matching decoder
    \item Make decisions based on pattern recognition (e.g., detecting congestion)
\end{itemize}
While efficient, this system operates primarily through statistical pattern matching and reconstruction. If an unusual scenario occurs (e.g., a new type of vehicle or unexpected road obstacle), the system may fail to properly encode, transmit, or interpret this information.

\subsubsection{System 2 SC: Future Directions}

In contrast, System 2 SC goes beyond System 1 SC in many ways. First, in terms of targeted use cases (e.g., human-machine collaboration), then in terms of its cognitive capabilities rooted in logical reasoning, planning, abstraction and analogy-making. Here, the needed notion of information  goes beyond Shannon information towards higher-order information structures and categories (see Section \ref{chamaeleon}). 
What is more? semantics is not just another buzzword but has precise mathematical foundations. While the details are beyond our scope, we can understand the key ideas through analogies:
\begin{itemize}
    \item \textbf{Categories and functors} can be thought of as systems of objects (like concepts) and the transformations between them. Just as we can translate from English to French while preserving meaning, functors map between different representational systems while preserving their structure.
    
    \item \textbf{Model theory} provides a framework for relating symbolic languages to the worlds they describe. A \textbf{model} is essentially an interpretation that makes statements in a language true or false. When two models have the same structure (a \textbf{homomorphism} between them), we can translate knowledge from one domain to another, like applying physics principles learned in a classroom to real-world engineering problems.
\end{itemize}
These mathematical tools allow us to precisely define how meaning is preserved across different representations and reasoning systems, which is a crucial capability for System 2 SC.\footnote{Categorical semantics provides a formal framework for understanding how syntax (the form of expressions) relates to semantics (their meaning). Natural transformations allow us to convert between different functors in a way that respects their structure (preserving meaning).}
System 2 SC is precisely what will enable human-robot interaction/collaboration, collaborative robots, remote teaching of skills and other sci-fi use cases. Aren't those the use cases 6G promised? Finally, it is worth underscoring that both systems will be needed, and it is their interaction that will be revolutionary. And just to make it clear, we still need Shannon information.  Pursuing System 2 SC requires going beyond statistical methods and probability theory, towards using algebraic topology, logic and category theory. Without this, semantic communication will remain incremental \& based on re-packaging existing works. Very early preliminary works going beyond System 1 SC can be found in \cite{10054510}, \cite{10296962}.

\textbf{Example 2: Image Transmission in System 2 SC.} For the same traffic monitoring scenario, a System 2 SC approach would operate fundamentally differently:
\begin{itemize}
    \item Extract structured information from the scene (e.g., ``three cars in left lane, one pedestrian crossing'')
    \item Maintain an ongoing world model that tracks object persistence and relationships
    \item Transmit only changes to the world model rather than raw or encoded pixels
    \item Reason about causality and potential futures (e.g., ``pedestrian is likely to reach other side in 5 seconds'')
    \item Adapt communication based on receiver's goals and knowledge state
\end{itemize}
This approach enables the system to handle novel situations by reasoning about them in terms of known concepts and their relationships, rather than relying solely on previously observed patterns.

\textbf{Example 3: Complementary role.} We can see the complementary nature of these systems in an autonomous vehicle scenario. System 1 SC handles routine perception and communication tasks efficiently (transmitting road conditions, processing sensor data, and executing learned driving patterns). Meanwhile, System 2 SC activates when the vehicle encounters unusual situations, reasoning about novel obstacles, communicating with human drivers using natural language, or planning alternative routes when faced with unexpected road closures. The vehicle might use System 1 for 95\% of its operations, but System 2 capabilities are essential for handling the remaining 5\% of edge cases that would otherwise lead to failures.

\subsection{Research Thrusts}\label{sec:research_thrusts}
This appendix outlines our proposed research agenda for realizing the vision of emergent reasoning-driven compositional systems. The \textit{four} synergistic research thrusts (RTs) address complementary aspects of our framework: RT1 focuses on representation learning and compositionality, RT2 develops the theoretical foundations for reasoning-driven semantic communication, RT3 applies these principles to emerge novel adaptive, resilient  communication protocols for practical applications, and RT4 bridges knowledge representation and reasoning through hierarchical world models and compositional inference mechanisms. Each thrust integrates statistical, topological, and logical perspectives of information (Sec. II), representations (abstractions), communication and protocols/languages  to ensure both theoretical rigor and practical utility.

\vspace{5pt}

\noindent\textbf{\textsf{RT1: Sheaf-theoretic Framework for  multimodal representation learning and Compositionality}} 
\vspace{+1mm}

RT1 proposes a novel \textbf{distributed sheaf-theoretic} framework for  \textbf{learning   multimodal  representations and how to algebraically compose them} for solving downstream tasks. Unlike traditional approaches that struggle with information heterogeneity, our sheaf-theoretic framework directly addresses the challenge of combining semantically compatible but structurally diverse representations. First,  nodes  learn  abstractions  from their  (multimodal) sensory signals via  interaction with  the environment and other agents. 
Subsequently, sheaf theory is used to \textit{compose} these  heterogeneous streams of semantic information (abstractions)  encoded in various algebraic data structures (vector spaces, lattices and topological spaces). 
The core idea is that, instead of studying the space itself, we study algebraic data structures (i.e., sheaves) and their transformations, which describes the space of semantic information where reasoning tasks, coordination and communication occur.  A preliminary work can be found  in \cite{issaid2025tacklingfeaturesampleheterogeneity}

\vspace{+2mm}

\noindent\textbf{\textsf{RT2: Emergent Reasoning-driven  Semantic Communication}}
\vspace{+1mm}

RT2 advances beyond statistical approaches to communication (System 1 SC) towards reasoning-driven semantic communication (System 2 SC, see Appendix~\ref{sec:sc12}). This thrust investigates how agents with different internal world models can establish effective communication through shared abstractions and reasoning processes. Unlike current approaches that assume homogeneous agents or predefined protocols, RT2 explores how meaningful communication emerges naturally from agents' need to coordinate and achieve shared goals.  RT2 addresses how and under what conditions  communication emerges among agents with partial information and limited knowledge,  while being robust to different  priors, beliefs and structures.  We demonstrate how topological/logical/causal abstractions confer robustness, structural semantics,  and interpretability of the emergent language.    
\vspace{+2mm}

\vspace{+1mm}
\noindent\textbf{\textsf{RT3:   Adaptive, Resilient  and Semantic Communication Protocols}}
\vspace{+1mm}

Learning communication protocols from data can transform both data,  control and knowledge planes. A medium access control (MAC) protocol defines rules for exchanging signaling messages to manage communication control actions, where  the goal is to learn and emerge vocabularies (syntax) and signaling for various traffic and service types. Current \textit{System 1 MAC } approaches are brittle, lack generalization, explainability,  scalability and cannot be deployed in mission/safety-critical applications. The
 proposed framework addresses fundamental challenges in ensuring adaptive, generalized, efficient and resilient communication protocols. Specifically,  RT3 proposes a data and reasoning-driven framework for learning communication protocols tailored to application requirements while taking into account resource constraints. RT3  examines the statistical, topological, and logical aspects of protocol learning, through the lens of \textbf{planning and reasoning with multiscale, multistep world models}.

\vspace{+1mm}
\noindent\textbf{\textsf{RT4: Hierarchical World Models and Compositional Inference Mechanisms}}
\vspace{+1mm}

RT4 focuses on developing scalable frameworks for learning \textbf{hierarchical world models and composable inference mechanisms} that enable effective reasoning in complex environments. While deep learning has succeeded in pattern recognition, it struggles with causal reasoning and counterfactual inference. We propose a principled separation between world models (knowledge representation) and inference mechanisms (reasoning processes) inspired by cognitive architectures \cite{lake2017building, battaglia2018relational, ha2018recurrent}. Hierarchical world models will capture multi-scale causal structures spanning from low-level sensorimotor patterns to high-level abstract concepts, enabling agents to reason across different levels of abstraction. For inference, we explore GFlowNets \cite{bengio2023gflownetfoundations} alongside other approaches like amortized variational inference, Monte Carlo tree search \cite{silver2018general}, and neuro-symbolic methods \cite{garcez2020neurosymbolic}. Unlike monolithic approaches, this modular design allows world models to be interpretable, compositional, and reusable across tasks, while inference mechanisms can be specialized and adapted to specific domains. The framework addresses fundamental challenges in compositional generalization \cite{lake2018generalization}, out-of-distribution robustness \cite{arjovsky2019invariant}, and sample efficiency by incorporating structural priors about causality \cite{pearl2009causality, scholkopf2021toward} and uncertainty into both representations and reasoning processes.
\vspace{+2mm}
\noindent

\end{document}